%% file: FLcriticalperiodsarXiv.tex
\documentclass{article}

\usepackage[square,sort,comma,numbers]{natbib}
\usepackage[preprint]{neurips_2020}

\usepackage[utf8]{inputenc} % allow utf-8 input
\usepackage[T1]{fontenc}    % use 8-bit T1 fonts
\usepackage{hyperref}       % hyperlinks
\usepackage{url}            % simple URL typesetting
\usepackage{booktabs}       % professional-quality tables
\usepackage{amsfonts}       % blackboard math symbols
\usepackage{nicefrac}       % compact symbols for 1/2, etc.
\usepackage{microtype}      % microtypography
\usepackage{wrapfig}
\usepackage{graphicx}
\usepackage{algorithm}
\usepackage{algorithmic}
\usepackage{color} 

\usepackage{amsmath,amsfonts,comment,bbm,amssymb}

\newcommand{\otoprule}{\midrule[\heavyrulewidth]}

\title{Critical Learning Periods in Federated Learning}

\author{
Gang Yan\\
SUNY-Binghamton University\\
Binghamton, NY 13902\\
 \texttt{gyan2@binghamton.edu}
   \And
Hao Wang\\
Louisiana State University\\ 
Baton Rouge, LA 70803\\
 \texttt{haowang@lsu.edu}
  \And
Jian Li\\
SUNY-Binghamton University\\ 
Binghamton, NY 13902\\
 \texttt{lij@binghamton.edu}
}

\begin{document}

\maketitle

\input{abstract}
\input{intro}

\input{prelim}

\input{exp}

\input{fisher}

\input{seizing}

\input{related}
\input{conclusion}

\bibliographystyle{unsrt}  
\bibliography{refs}

\clearpage
\appendix
\input{appendix}

\end{document}

%% file: abstract.tex
\begin{abstract}
Federated learning (FL) is a popular technique to train machine learning (ML) models with decentralized data.  Extensive works have studied the performance of the global model; however, it is still unclear how the training process affects the final test accuracy.  Exacerbating this problem is the fact that FL executions differ significantly from traditional ML with heterogeneous data characteristics across clients, involving more hyperparameters. 
In this work, we show that the final test accuracy of FL is dramatically affected by the early phase of the training process, i.e., FL exhibits critical learning periods, in which small gradient errors can have irrecoverable impact on the final test accuracy.  To further explain this phenomenon, we generalize the trace of the Fisher Information Matrix (FIM) to FL and define a new notion called FedFIM, a quantity reflecting the local curvature of each clients from the beginning of the training in FL.  Our findings suggest that the {\em initial learning phase} plays a critical role in understanding the FL performance.  This is in contrast to many existing works which generally do not connect the final accuracy of FL to the early phase training.  Finally, seizing critical learning periods in FL is of independent interest and could be useful for other problems such as the choices of hyperparameters such as the number of client selected per round, batch size, and more, so as to improve the performance of FL training and testing.

\end{abstract}

%% file: intro.tex
\section{Introduction}\label{sec:intro}

The ever-growing attention to data privacy and the popularity of mobile computing have impelled the rise of Federated learning (FL)~\cite{mcmahan2017communication,kairouz2019advances,imteaj2021survey}, a new distributed machine learning paradigm on decentralized data.
A typical FL system consists a central server and multiple decentralized clients (e.g., smartphones and IoT devices). The central server initiates federated learning by sending a global model to clients. The clients then use their local data samples to train the received model with common deep learning algorithms and aggregate their local models to the central server. The central server updates the global model by aggregating the received local models and sends it to clients for further training. By repeating the local training and global aggregation, the central server obtains a global model jointly trained by decentralized clients without leaking any raw data. This unique distributed nature enables an extensive deployment of FL that trains deep learning models on sensitive private data, such as Google Keyboard~\cite{yang2018applied}.

The distributed nature of FL raises a series of new challenges in terms of \textit{system performance} and \textit{data statistics}. In FL systems, clients are typically loosely-connected mobile devices with limited communication bandwidth, computation power, and battery life. Besides, unlike traditional centralized machine learning (ML), data samples of each client in FL follow a non-identical and independent distribution (non-IID), introducing bias that slows down or even fails the training. 
A few recent studies have been proposed to address these challenges by model compression~\cite{konevcny2016federated,suresh2017distributed,caldas2018expanding,xu2019elfish}, communication frequency optimization~\cite{wang2019adaptivecom,wang2019adaptive,karimireddy2020scaffold}, and client selections~\cite{lai2021oort, wang2020optimizing,xiong2021straggler}.

However, existing studies have not yet explored the significance of {\em critical learning periods}. 
Recent works have revealed that the first few training epochs---known as critical learning periods---determine the final quality of a deep neural network (DNN) model in traditional centralized ML~\cite{achille2019critical,jastrzebski2019relation,golatkar2019time,jastrzebski2021catastrophic}. During a critical period, deficits such as low quality or quantity of training data will cause irreversible model degradation, no matter how much additional training is performed after the period.  The existence of critical periods in FL remains an open question due to the unique distributed nature of FL.

In this paper, we seek critical learning periods in FL with systematic experiments and theoretical analysis, and we emphasize the necessity of seizing the critical learning periods to improve FL training efficiency. Specifically, through a range of carefully designed experiments on different ML models and datasets, we observe the consistent existence of critical learning periods in the FL training process.  We further propose a new metric named Federated Fisher Information Matrix (FedFIM) to describe and explain this phenomenon. FedFIM is calculated based on a classical statistics notion of Fisher Information Matrix (FIM)~\cite{amari2000methods} that approximates the local curvature of the loss surface in FL efficiently. 
We show that the phenomenon of critical learning periods in FL can be explained using the trace of FedFIM, a quantity reflecting the local curvature of each clients from the beginning of the training in FL.  Our findings suggest that the {\em initial learning phase} plays a critical role in understanding the FL performance, complementing many existing studies that generally ignore the connection between the final model accuracy and the early phase training.  To the best of our knowledge, this is the first work towards seizing critical learning periods in FL framework for training efficiency.
Our main contributions are as follows: 
\begin{enumerate}
\item We discover that critical learning periods consistently exist in FL with representative models and datasets through our carefully-designed experiments.
\item We systematically explore the impacts of critical learning periods for FL under a wide range of FL hyperparameters, including client availability, learning rates, batch size, and weight decay. 
\item We propose a new notion dubbed Federated Fisher Information Matrix (FedFIM) and analyze the phenomenon of critical learning periods in FL through the trace of FedFIM. We show that model quality during critical periods correlates strongly with the trace of FedFIM. 
\end{enumerate}

%% file: prelim.tex
\section{Background}

\subsection{Federated Learning}

The goal of FL is to solve a joint optimization problem as
\begin{align}\label{eq:opt-fl}
\min_{\textbf{w}\in\mathbb{R}^d} \mathcal{L}(\textbf{w}, \mathcal{D}):= \sum_{j\in\mathcal{N}} \frac{|\mathcal{D}_j|}{|\mathcal{D}|} \mathcal{L}_j(\textbf{w},\mathcal{D}_j),
\end{align}
where $\textbf{w}$ denotes the model parameters, $\mathcal{N}$ denotes the set of clients, $\mathcal{D}_j$ is the local dataset of client $j\in\mathcal{N}$, the entire training dataset is $\mathcal{D}=\cup_{j\in\mathcal{N}}\mathcal{D}_j$, and $\mathcal{L}_j(\textbf{w},\mathcal{D}_j)$ is the local loss function of client $j$.  A typical solution to this optimization problem is \textit{federated averaging} (FedAvg) algorithm \cite{mcmahan2017communication}. Specifically, FedAvg initializes with a random global model $\textbf{w}_0$ and iterates the following two steps within each communication round $t=1,\cdots, T$:
\begin{itemize}
\item \textbf{Local training.} The central server sends the goal model $\textbf{w}_{t-1}$ to a randomly selected subset of clients $\mathcal{N}_t\subset\mathcal{N}$.  Each selected client $j\in\mathcal{N}_t$ performs local training using its own dataset $\mathcal{D}_j$:
\begin{align}\label{eq:local-update}
\textbf{w}_{t, j}(k)\leftarrow \textbf{w}_{t, j}(k-1)-\eta \nabla \mathcal{L}_j(\textbf{w}_{t, j}(k-1),\mathcal{D}_j),
\end{align}
where $\eta$ is the learning rate and $k=1,\cdots,K$ is the index of local iterations.  
\item \textbf{Global aggregation.} The central server obtains a new global model $\textbf{w}_t$ by weighted-averaging the local models collected from the selected clients in round $t$:
\begin{align}\label{eq:global}
\textbf{w}_t \leftarrow \sum_{j\in\mathcal{N}_t} \frac{|\mathcal{D}_j|}{|\cup_{j\in\mathcal{N}_t}\mathcal{D}_j|}\textbf{w}_{t, j}(K). 
\end{align} 
\end{itemize}

There are a few variant federated learning algorithms, such as SCAFFOLD~\cite{karimireddy2020scaffold}, FedProx~\cite{li2020federated}, and FetchSGD~\cite{rothchild2020fetchsgd}. We choose to perform observations and analysis based on FedAvg because its simplicity and generality extensively reduce the uncertainty of critical periods.

\subsection{Critical Learning Periods}

Critical learning periods were originally observed in early post-natal development of humans and animals that sensory deficits will cause lifelong irreversible skill impairment. Recently, researchers observed similar phenomenons in centralized deep learning that training a model with defective data such as blurred images in early epochs will decrease its final accuracy, no matter how many additional training epochs are performed~\cite{agarwal2020accordion,achille2019critical,jastrzebski2019relation,golatkar2019time,jastrzebski2021catastrophic}. 

However, observing and justifying critical learning periods in FL are hindered a few obstacles: (i) FL involves multiple deep learning processes across randomly selected clients with their own data; (ii) the global model aggregated by local models at the central server has no direct information about the training data decentralized across clients; and (iii) FL has far more hyperparameters (e.g., the number of selected clients and data distribution) than centralized training that make it complicated to induce critical learning periods.

%% file: exp.tex
\begin{figure*}[t]
	\center
    	\includegraphics[width=0.95\textwidth]{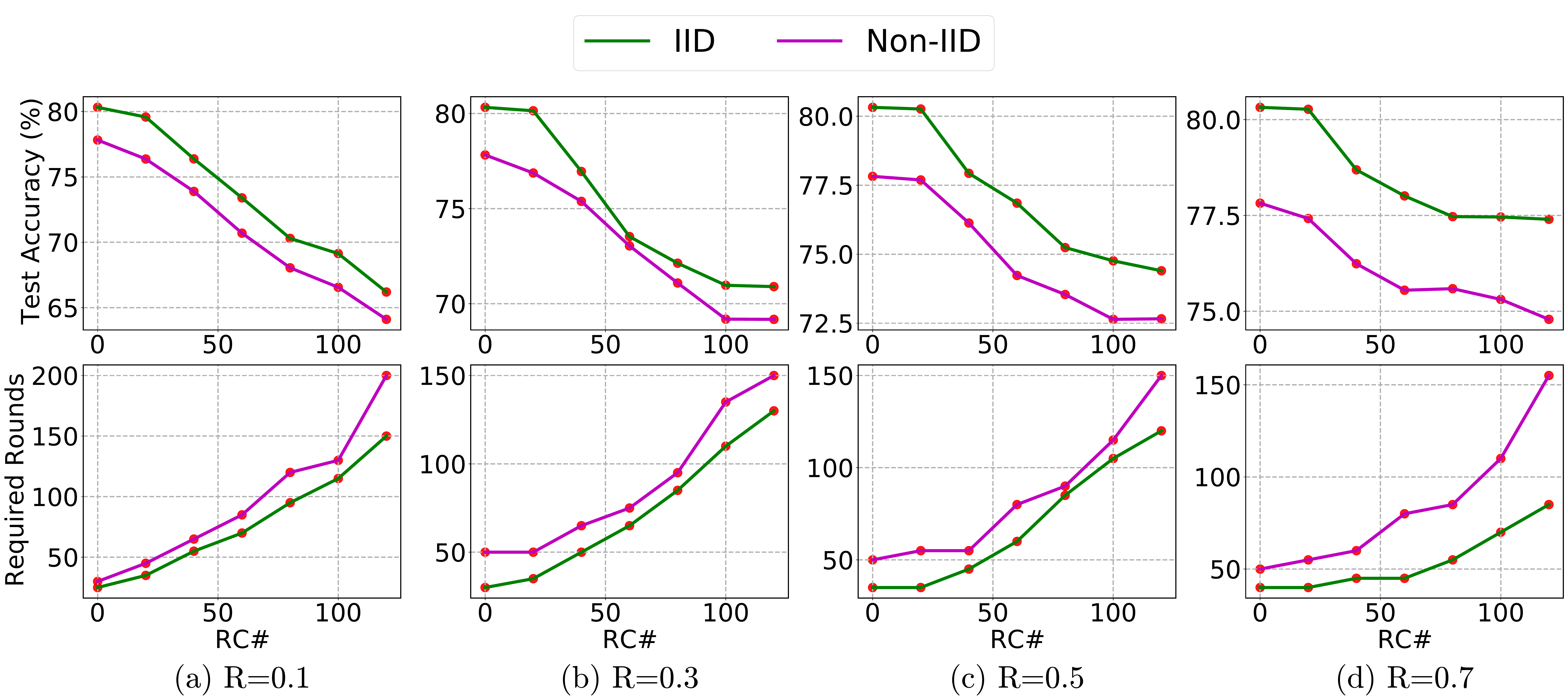}
		\caption{FL exhibits critical learning periods.  (top) The final accuracy achieved by \textbf{ResNet-18} on both IID and Non-IID \textbf{CIFAR-10} with FedAvg using partial local datasets (where $R$ indicates the ratio of local datasets) for training as a function of the communication round at which the partial training dataset is recovered to the entire training dataset. The test accuracy of FL is permanently impaired if the training dataset is not recovered to the entire training dataset early enough, no matter how many additional training rounds are performed. (bottom) Communication rounds vs. recover round (RC\#).  The total communication rounds required to achieve the corresponding final accuracy are significantly increased as a function of the recover round. }
	\label{fig:acc-vs-recovery-all}
\end{figure*}

\section{Critical Learning Periods in FL}\label{sec:existence}

We hypothesize that the final accuracy of FL is significantly affected by the initial learning phase, which we term as the critical learning periods in FL.  Consider a model with loss function $\ell(\boldsymbol x; \textbf{w})$, where $\ell$ reaches a minimum loss $\ell_{\text{loss}}$ with a test accuracy $\ell_{\text{acc}}$ when optimized with FedAvg across $N$ decentralized clients on the entire training dataset $\mathcal{D}$. In addition, consider optimizing FedAvg across all clients only with a subset of the local training dataset $\mathcal{D}^\prime_j\subset\mathcal{D}_j$, $\forall j\in\mathcal{N}$ in the first $M$ communication rounds and then using the entire training dataset $\mathcal{D}$ afterwards.  Then $\ell$ reaches a minimum loss of $\ell^\prime_{\text{loss}}(M)$ with a test accuracy of $\ell^\prime_{\text{acc}}(M)$.  The critical learning periods articulate that there exist $M_1$ and $M_2$ such that $\ell^\prime_{\text{acc}}(M_1)\geq\ell^\prime_{\text{acc}}(M_2)$ when $M_1\leq M_2$, i.e., the initial learning phase is critical in determining the final performance of FL, and the effect of insufficient training (i.e., only using part of the entire training dataset) during the critical learning periods cannot be overcome, no matter how much additional training is performed.  

In this section, we address two key questions pertains to the phenomenon of critical learning periods in FL.  We first show via an extensive set of experiments that the critical learning periods can be observed across different popular ML models and datasets.  We then reveal that the critical learning periods in FL stay robust under various training schemes.

\begin{wrapfigure}{rt}{0.5\linewidth}
    \centering
      \includegraphics[width=0.5\textwidth]{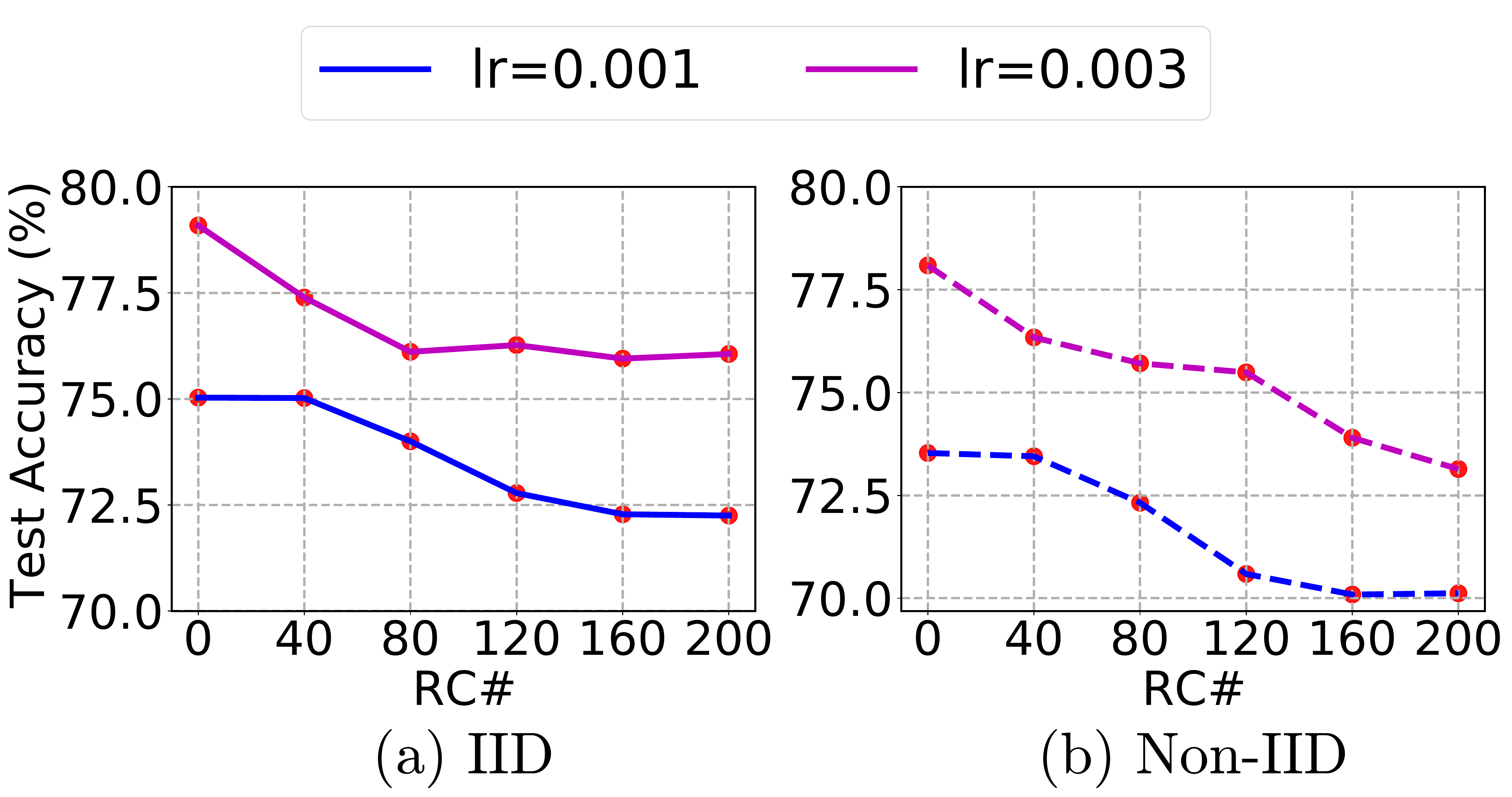}
      \vspace{-0.2in}
		\caption{The existence of critical learning periods in FL: FedAvg trained on \textbf{ResNet-18} using both IID and Non-IID \textbf{CIFAR-10} with constant learning rates (lr). }
	\label{fig:fixed-learning-rate}
\end{wrapfigure}

\subsection{FL Exhibits Critical Learning Periods}
We perform extensive simulations using two representative ML models: ResNet-18 \cite{he2016deep} and CNN, on popular datasets CIFAR-10 and CIFAR-100 \cite{krizhevsky2009learning}.  To present the existence of critical learning periods in FL, we adopt the standard FedAvg \cite{mcmahan2017communication} which requires the entire training dataset throughout the training process, as well as its performance when only a subset of the training dataset on each client is involved in the first $M$ communication rounds at which the training dataset is recovered to the entire training dataset.  We call $M$ as the ``Recover Round'' and denote $R$ as the ratio of local datasets involved in training.  We consider a system with $N=64$ clients and FedAvg randomly selects a subset of $12$ clients in each round.  The batch size is of $16$; the initial learning rate is set to $0.01$ with a decay of $0.97$ per round; and the SGD solver is adopted using an exponential annealing scheduling for the learning rate with a weight decay of $5\times 10^{-4}$.

Figure~\ref{fig:acc-vs-recovery-all} (top) reports the final performance of FL affected by the partial training datasets with different ratios $R$ as a function of the recover round $M$.  All results consistently endorse that the critical learning periods exist across all settings with different ratios of local datasets involved in the early learning phase: if the training dataset is not recovered to the entire dataset, at as early as the $20$-th communication rounds, the final test accuracy of FL is severely degraded compared to the standard FedAvg.  Comparing among different ratios $R$ of local datasets involved in early training phase, it is not too surprising to see that lower $R$ of local datasets in the early training phase makes drawing critical learning periods easier.  

We further measure the total communication rounds required to achieve the corresponding final accuracy as a function of the recover round, as illustrated in  Figure~\ref{fig:acc-vs-recovery-all} (bottom).
It is obvious that the communication rounds are significantly increased with a lower final test accuracy as a function of the recover round $M$.  This further indicates the importance of the initial learning phase in determining the final performance of FL. 

\begin{wrapfigure}{rt}{0.5\linewidth}
    \centering
    	\includegraphics[width=0.5\textwidth]{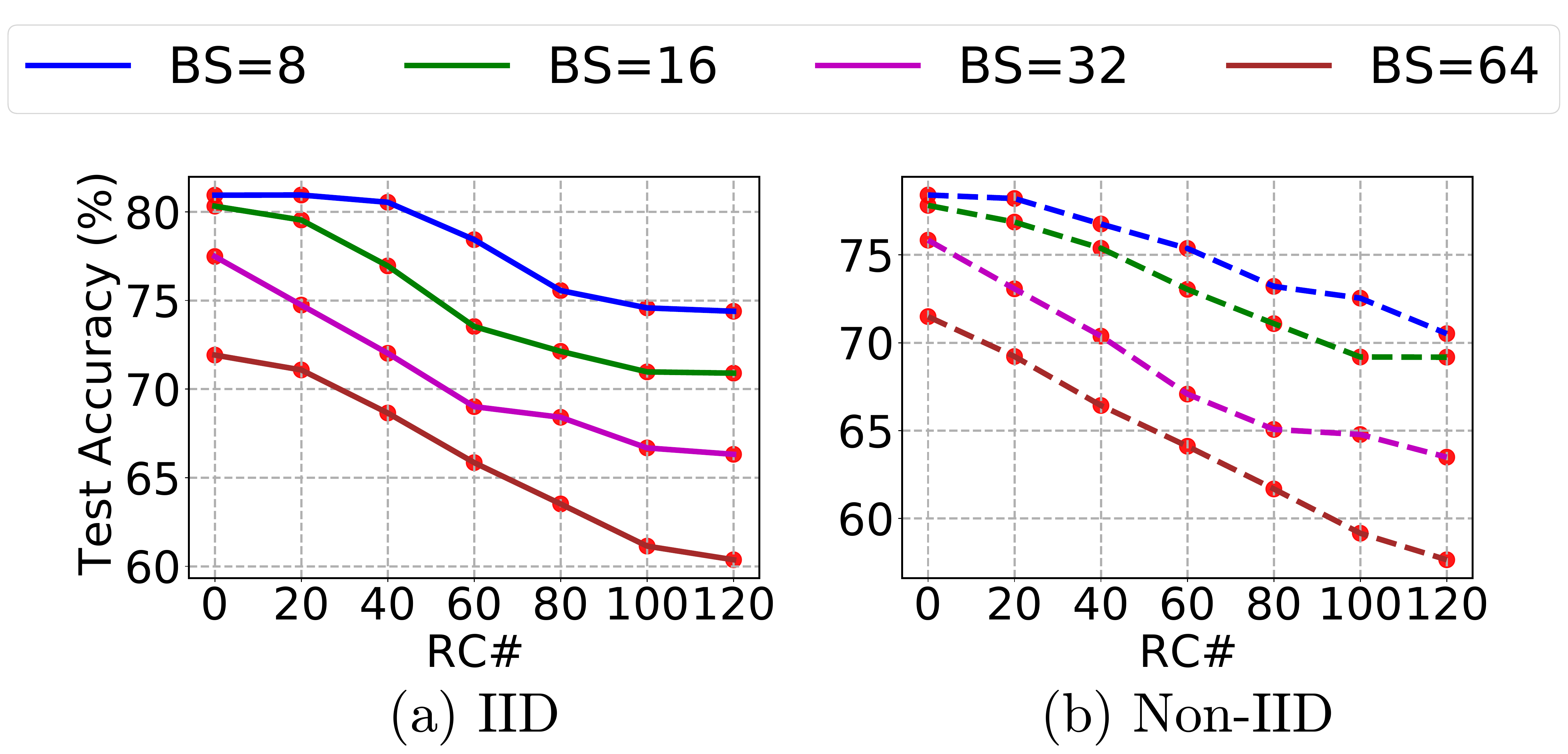}
    	\vspace{-0.2in}
		\caption{The existence of critical learning periods in FL: FedAvg trained on \textbf{ResNet-18} using both IID and Non-IID \textbf{CIFAR-10} with different batch sizes (BS).}
	\label{fig:batch-size}
\end{wrapfigure}

\subsection{Learning Rate Annealing and Batch Size} 
We conduct the same experiments as in Figure~\ref{fig:acc-vs-recovery-all} but using a constant learning rate rather than an annealing scheme.  In particular, we set the constant learning rates to be $0.001$ and $0.003$, respectively.  From Figure~\ref{fig:fixed-learning-rate}, we still observe the existence of critical learning periods in FL even with constant learning rates.  Therefore the phenomenon of critical learning periods in FL are not resultant from an annealed learning rate in later rounds, and cannot be solely explained in terms of the loss landscape of the optimization in~(\ref{eq:opt-fl}).  
Analogous results illustrating the impact of batch size are presents in Figure~\ref{fig:batch-size}. Again the critical learning periods consistently exist regardless of the choice of batch size.  This further suggests that the phenomenon of critical learning periods in FL cannot be simply explained by the differences in batch sizes.

\begin{wrapfigure}{rt}{0.5\linewidth}
    \centering
    \vspace{-0.2in}
    	\includegraphics[width=0.5\textwidth]{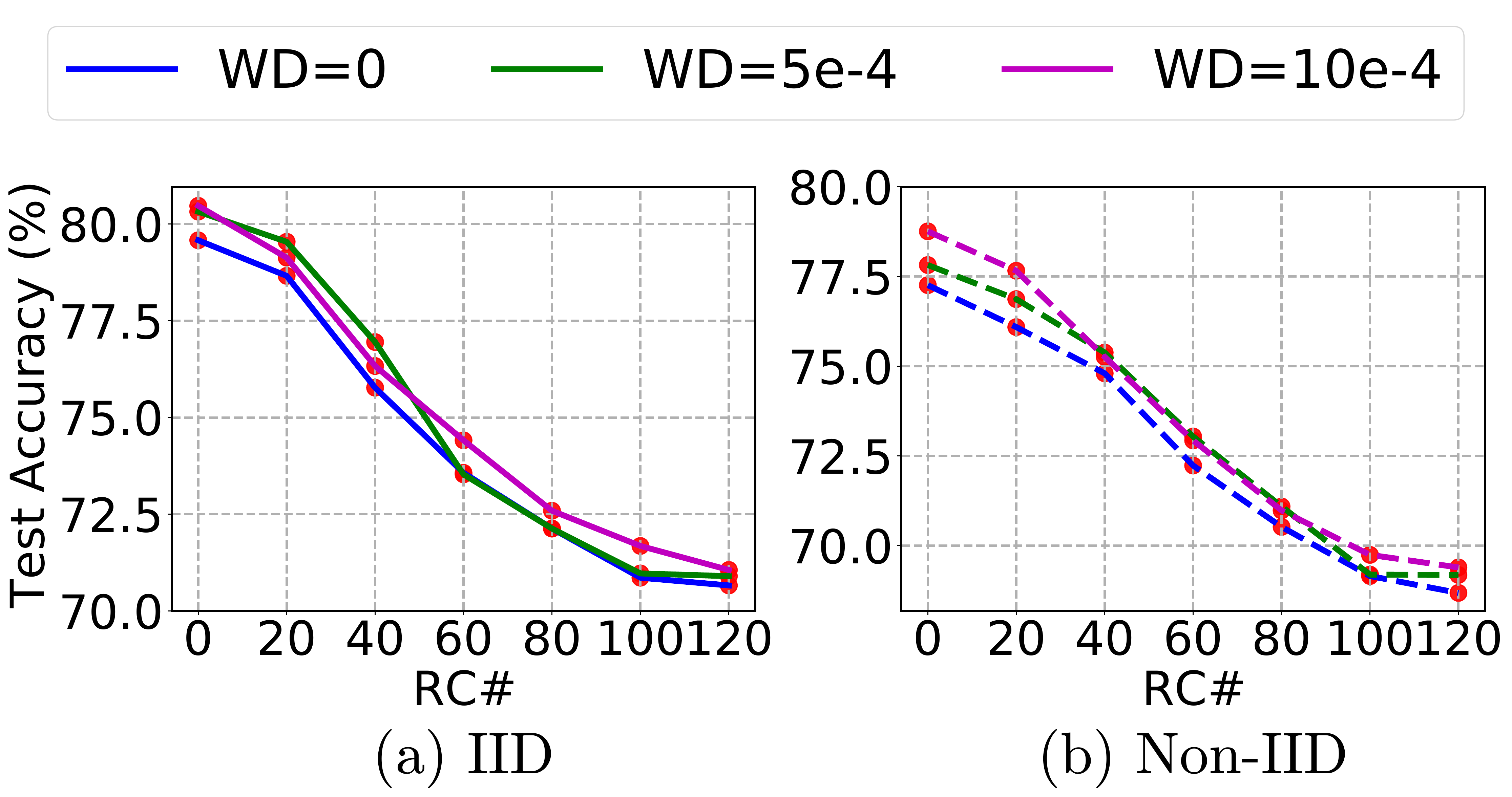}
    	\vspace{-0.2in}
		\caption{The existence of critical learning periods in FL: FedAvg trained on \textbf{ResNet-18} using both IID and Non-IID \textbf{CIFAR-10} with different weight decays (WD).}
	\label{fig:weight-decay}
		\vspace{-0.4in}
\end{wrapfigure}

\subsection{Weight Decay}
Similarly, the results for the same experiments as in Figure~\ref{fig:acc-vs-recovery-all} but with different weight decays are presented in Figure~\ref{fig:weight-decay}. We still observe the critical learning periods as in Figure~\ref{fig:acc-vs-recovery-all}, but surprisingly the shapes of the critical learning periods are robust to the values of weight decays, i.e., changing the weight decays does not impact the shape of the critical learning periods.

%% file: fisher.tex
\section{Federated Fisher Information}\label{sec:fim}

Through extensive experiments, we have shown that the \textit{initial learning phase} of the training process plays a critical role in the final test accuracy of FL.  Our main contribution in this section is to show that this phenomenon can be explained by the trace of the \textit{Federated Fisher Information Matrix (FedFIM)}, a quantity reflecting the local curvature of each clients from the beginning of the training in FL.   We begin with the definition of the FIM for centralized training.

\subsection{Fisher Information Matrix}
Consider a probabilisitic classification model $p_{\textbf{w}}(y|\boldsymbol x)$, where $\textbf{w}$ is the model parameter.  Let $\ell(\boldsymbol x, y;\textbf{w})$ be the cross-entropy loss function calculated for input $\boldsymbol x$ and label $y.$ Denote the corresponding gradient of the loss computed for an example $(\boldsymbol x, y)$ as $g(\boldsymbol x, y;\textbf{w})=\frac{\partial}{\partial\textbf{w}}\ell(\boldsymbol x, y;\textbf{w})$.  Then the Fisher Information Matrix (FIM) $\boldsymbol F$ for centralized training is defined as 
\begin{align}\label{eq:fim}
\boldsymbol F(\textbf{w})=\mathbb{E}_{\boldsymbol x\sim \mathcal{X}, \hat{y}\sim p_{\textbf{w}}(y|\boldsymbol x)}[g(\boldsymbol x, \hat{y})g(\boldsymbol x, \hat{y})^\intercal],
\end{align}
where the expectation is often approximated using the empirical distribution $\mathcal{X}$ induced by the centralized training dataset.  Note that the FIM can be viewed as a local metric on how much the perturbation of the weights affects the network output \cite{amari2000methods}.  The FIM can also be seen as an approximation to the Hessian of the loss function \cite{martens2014new}, and hence of the curvature of the loss landscape at a particular point $\textbf{w}$ during training.  This provides a natural connection between the FIM and the optimization procedure \cite{amari2000methods}.  

However, the computation of FIM in~(\ref{eq:fim}) requires the availability of the entire training dataset for the global model at the server.  Unfortunately, this is infeasible for FL since training data is decentralized across clients.  Hence we cannot compute FIM for FL as in~(\ref{eq:fim}).  We now introduce a new notion to overcome this challenge.  

\begin{figure*}
	\center
    	\includegraphics[width=0.99\textwidth]{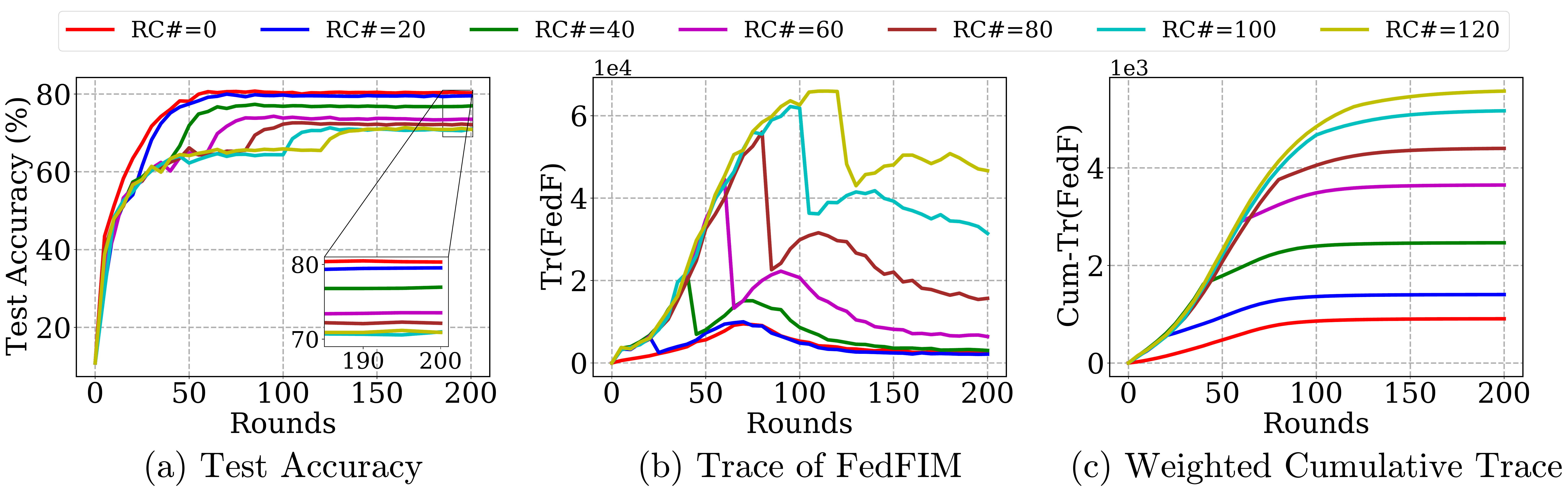}
		\caption{Connections between critical learning periods in FL and the Federated Fisher information achieved by \textbf{ResNet-18} on IID \textbf{CIFAR-10} with FedAvg using 30\% of local datasets for training initially and recover to the entire datasets upon the recover round. (a) Test accuracy vs. recover rounds: the final test accuracy is permanently impaired if the training dataset is not fully recovered at as early as the 20-th round. (b) Trace of FedFIM vs. recover round. There exists a sharp increase of the trace of FedFIM in the early training phases. (c) Weighted cumulative sum of the trace of FedFIM vs. recover round. }
	\label{fig:fim-iid}
\end{figure*}

\begin{figure*}
	\center
    	\includegraphics[width=0.99\textwidth]{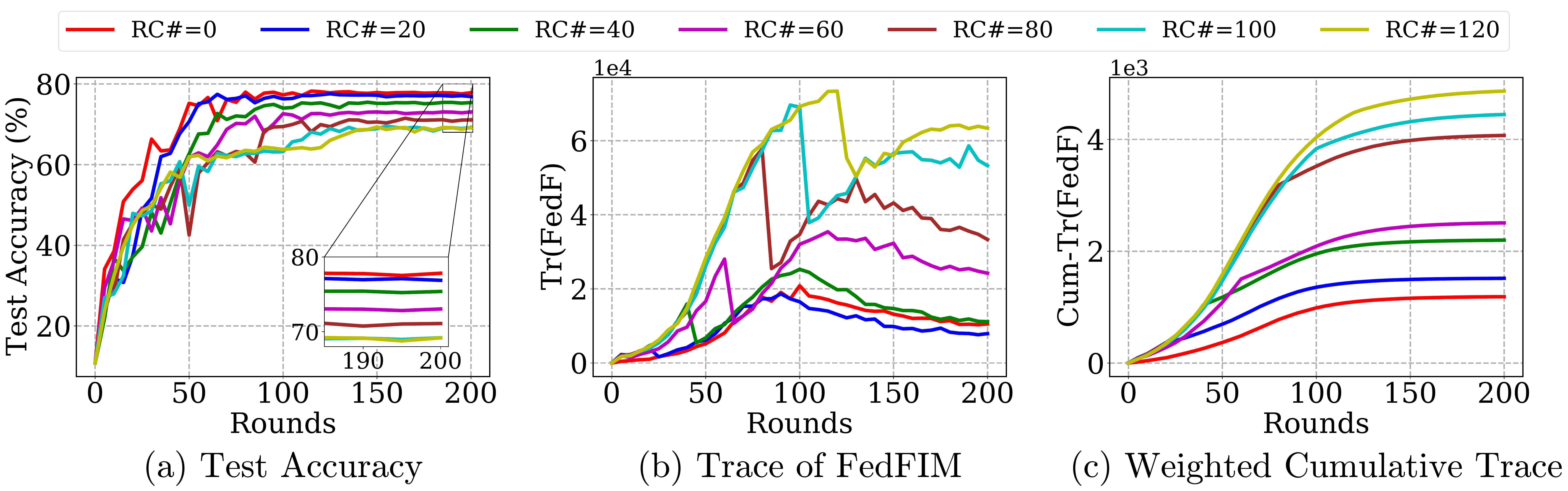}
		\caption{Connections between critical learning periods in FL and the Federated Fisher information achieved by \textbf{ResNet-18} on Non-IID \textbf{CIFAR-10} with FedAvg using 30\% of local datasets for training initially and recover to the entire datasets upon the recover round.}
	\label{fig:fim-non-iid}
	\vspace{-0.1in}
\end{figure*}

\subsection{Federated Fisher Information Matrix}  
Given that training data resides in each client, and the training process of FL in~(\ref{eq:local-update}) and~(\ref{eq:global}), we first introduce the notation of $\boldsymbol F_j(\textbf{w})$, which represents the local FIM on client $j\in\mathcal{N}$:  
\begin{align}\label{eq:fim-client}
\boldsymbol F_j(\textbf{w})=\mathbb{E}_{\boldsymbol x_j\sim \mathcal{X}_j, \hat{y}_j\sim p_{\textbf{w}}(y_j|\boldsymbol x_j)}[g(\boldsymbol x_j, \hat{y}_j)g(\boldsymbol x_j, \hat{y}_j)^\intercal],
\end{align}
where $\mathcal{X}_j$ is the empirical distribution induced by the local dataset $\mathcal{D}_j$ of client $j$.  Note that $\boldsymbol F_j(\textbf{w})$ is computed using the global model $\textbf{w}$ on the local dataset $\mathcal{D}_j$, and can be considered as a local metric measuring how the perturbation of the global model affects the FL training performance from the perspective of client $j.$ As a result, the overall impact of the perturbation of the global model on the final output, which we define as the Federated Fisher Information Matrix (FedFIM) $\boldsymbol {FedF}$ for FL, can be computed using the weighted average of local FIM across all clients: 
\begin{align}\label{eq:fedfim}
\boldsymbol {FedF}(\textbf{w})=\sum_j \frac{|\mathcal{D}_j|}{|\mathcal{D}|} \boldsymbol F_j(\textbf{w}),
\end{align}
where the weight of client $j$ is the size of its dataset.  The rationale is that lower local FIM often has little effect on the final performance.  
We denote the trace of $\boldsymbol {FedF}$ as $\text{Tr}(\boldsymbol {FedF})$.

\subsection{Experimental Results} We conduct similar experiments as in Figure~\ref{fig:acc-vs-recovery-all} with partial local datasets involved in the initial learning phases and the training datasets recover to the entire datasets at the ``recover rounds'' (RC\#).  The test accuracy and the trace of FedFIM with different recover rounds and $R=0.3$ on IID and Non-IID CIFAR-10 are presented in Figures~\ref{fig:fim-iid} and~\ref{fig:fim-non-iid}.  

First of all, we again observe the existence of critical learning periods since if the training dataset is not recovered to the entire datasets, e.g., at as early as the 20-th communication rounds, the final test accuracy of FL is permanently impaired.  Second, this information is fully reflected via the trace of FedFIM as shown in Figure~\ref{fig:fim-iid} (b) for IID case and Figure~\ref{fig:fim-non-iid} (b) for Non-IID case.  We observe a sharp increase in the trace of the FedFIM in the early phases of the FL training process, which coincides with dramastic increase of the test accuracy in the early training phase.  The information starts to decrease when the test accuracy starts to plateau.  Since the training datasets are recovered from $30\%$ of local datasets to the entire datasets at the recover rounds, additional data further boosts the test accuracy as shown in Figure~\ref{fig:fim-iid} (a).  However, such a test accuracy boost decreases significantly as the recover rounds increase.  This further suggests that the initial learning phases play a critical role in the FL performance and the permanent model degradation is irreversible no matter how much additional training is performed after the critical learning periods.  Correspondingly, the accuracy boosting results in a slight increase in the trace of FedFIM, and the information decreases again when the test accuracy starts to plateau.

In general, the measures of test accuracy and trace of FedFIM are noisy, especially with Non-IID dataset as shown in Figure~\ref{fig:fim-non-iid}.  This is because for instance the learning rate has to be adjusted in order to compensate for possible generalization issues of the training process \cite{jastrzkebski2017three,smith2018don}. To this end, we further consider a weighted cumulative sum of the trace of FedFIM as follows
\begin{align}
    \text{Cum-Tr}(\boldsymbol {FedF})(k)=\sum\limits_{i=0}^k \eta_i \text{Tr}(\boldsymbol {FedF}_i),
\end{align}
where $\eta_i$ is the learning rate at the $i$-th round, and $\boldsymbol {FedF}_i$ is the Federated Fisher Information Matrix at the $i$-th round.  The trace of FedFIM represents the degree of whether the local data is good enough to improve the model. A larger values correspond to less model information. This is exactly observed in Figure~\ref{fig:fim-iid} (c) and Figure~\ref{fig:fim-non-iid} (c), where a late recovery results in larger weight cumulative trace.

%% file: seizing.tex
\section{Seizing Critical Learning Periods}\label{sec:seizing}

We use carefully-crafted experiments to evaluate the idea that seizes critical learning periods to improve the FL training efficiency, though existing literature largely ignore the critical learning periods in FL training process. 
The experiments run on PyTorch on Python 3 with NVIDIA RTX 3060 GPU.  The total number of clients is $64$ and a subset of $25$ clients are randomly selected in each round.

\begin{figure*}[t]
	\center
    	\includegraphics[width=1\textwidth]{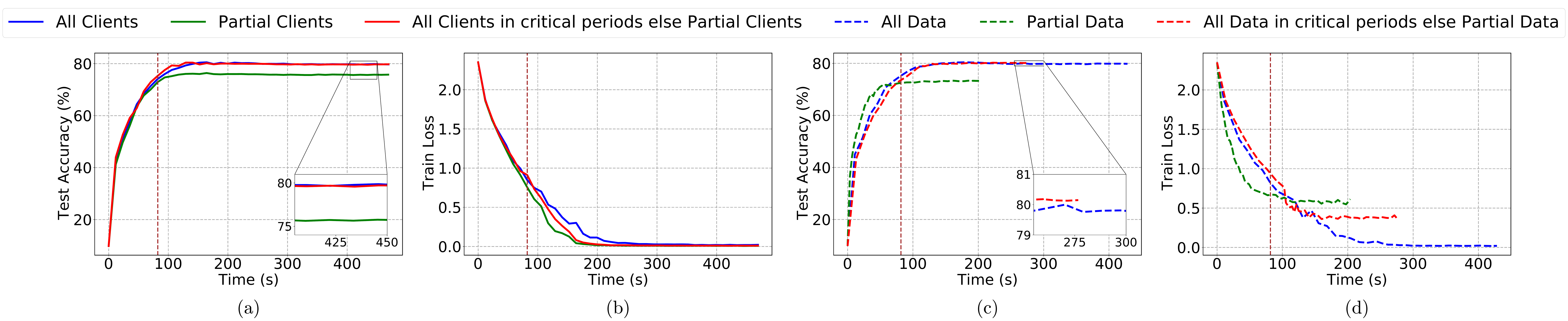}
    	\vspace{-0.2in}
		\caption{Seizing the critical learning periods in FL training with \textbf{ResNet-18} on IID \textbf{CIFAR-10}.}
	\label{fig:motivation-iid}
\end{figure*}

\begin{figure*}[t]
	\center
    	\includegraphics[width=1\textwidth]{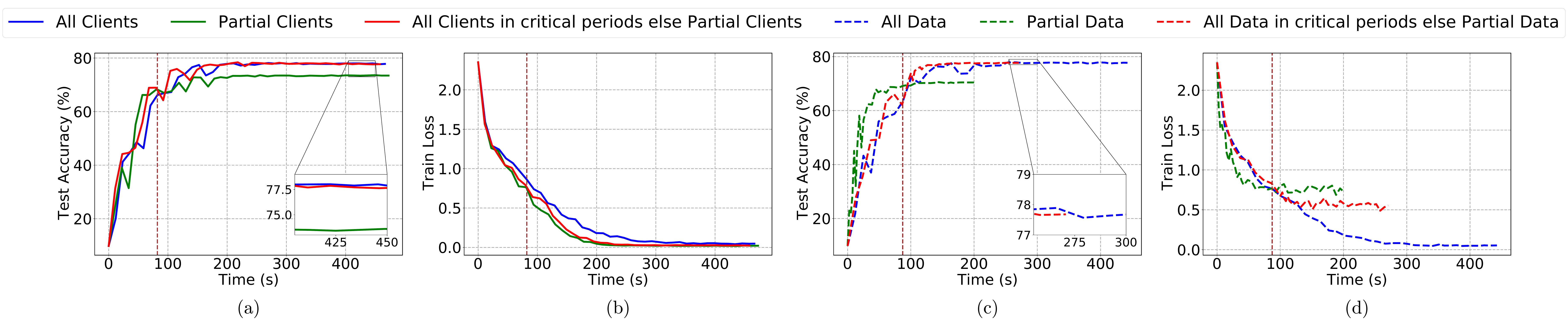}
    	\vspace{-0.2in}
		\caption{Seizing the critical learning periods in FL training with \textbf{ResNet-18} on Non-IID \textbf{CIFAR-10}.}
	\label{fig:motivation-non-iid}
	\vspace{-0.1in}
\end{figure*}

Specifically, we train ResNet-18 on IID and Non-IID CIFAR-10 with FedAvg under different settings as shown in Figures~\ref{fig:motivation-iid} and~\ref{fig:motivation-non-iid}:  
\begin{itemize}
    \item \textbf{All Clients}: All clients participate in federated learning.
    \item \textbf{Partial Clients}: Only a subset of the clients (e.g., $60\%$) participate in federated learning.
    \item \textbf{All Clients in critical periods else Partial Clients}: All clients participate in training during the critical learning periods. After that, only a subset of clients (e.g., $60\%$) remain in training. 
    \item \textbf{All Data}: Each client processes all data in local training.
    \item \textbf{Partial Data}: Each client processes only partial local datasets (e.g., $25\%$) in local training.
    \item \textbf{All Clients in critical periods else Partial Clients}: Each client uses its entire local dataset for training during the critical learning periods, and only uses their partial local dataset afterwards.
\end{itemize}

By seizing the critical periods in FL, we summarize the counter-intuitive experimental results as follows:

\textbf{No need to involve all clients in training all along.} The conventional FedAvg requires the entire training datasets across all clients throughout the training process.  However, some clients may not be available for training, e.g., due to unreliable network connection.  To illustrate the impact of critical learning periods, we further consider a heuristic in which all clients are involved in the training during the critical learning periods and then only a subset of clients (e.g., $60\%$) are involved afterwards.  

Figure~\ref{fig:motivation-iid}(a) and Figure~\ref{fig:motivation-non-iid}(a) show the test accuracy v.s. wall-clock time. There exists a requirement on the number of clients involved in training which provides similar test accuracy as using all clients (FedAvg) throughput.  For example, with all clients participate in the FL training during the critical learning periods, and then only 60\% of clients afterwards, the final test accuracy is similar to that using all clients throughout the training process.  Hence there is no need to involve all clients throughout the training process.  Figure~\ref{fig:motivation-iid}(b) and Figure~\ref{fig:motivation-non-iid}(b) show the train loss v.s. wall-clock time. The participated client number requirement reduces the training time than using all clients (FedAvg) throughput. It is clear that leveraging critical learning periods for FL training, even in a heuristic manner, can significantly improve the training efficiency with a reduced training time while maintaining final test accuracy.

\textbf{No need to train a model with all local data for each client.}
We consider the challenge that FL clients have heterogeneous system capabilities, e.g., can only process part of the local data for training.  We use a heuristic with entire local datasets used for training during the critical learning periods and then only partial local datasets involved afterwards.

Figure~\ref{fig:motivation-iid}(c) and Figure~\ref{fig:motivation-non-iid}(c) show the test Accuracy v.s. wall-clock time. There exists a training dataset requirement which provides similar test accuracy as using the entire dataset (FedAvg) throughput.  For example, with the entire training datasets used in the FL training during the critical learning periods, and then only 25\% of local datasets afterwards, the final test accuracy is similar to that using the entire datasets throughout the training process. Hence there is no need to use the entire training datasets throughout the training process.  
Figure~\ref{fig:motivation-iid}(d) and Figure~\ref{fig:motivation-non-iid}(d) present the train loss v.s. wall-clock time, the training dataset requirement (the heuristic) reduces the training time than using the entire dataset (FedAvg) throughput.  Again, we observe that the early learning phase plays a critical role in FL performance and leveraging it can significantly improve the training efficiency of FL.

Overall, we can save 40\%-50\% of the training time and 50\%-65\% of the total clients but achieve a close final model accuracy when training ResNet-18 on the IID and non-IID CIFAR-10 dataset.

%% file: related.tex
\section{Related Work}

Since the term of federated learning was introduced in the seminal work \cite{mcmahan2017communication}, there is an explosive growth in federated learning research.  For example, a line of works focuses on designing algorithms to achieve higher learning accuracy and analyze their convergence performance, e.g., \cite{smith2017federated,li2020convergence,liu2020federated,wang2020federated}.  Another line of works aim to improve the communication efficiency between the central server and clients through compressions or sparsification \cite{konevcny2016federated,suresh2017distributed,caldas2018expanding,xu2019elfish}, communication frequency optimization \cite{wang2019adaptivecom,wang2019adaptive,karimireddy2020scaffold}, client selections \cite{lai2021oort, wang2020optimizing,xiong2021straggler}, etc.  Additionally, a lot of efforts have been put on exploring the privacy and fairness of federated learning \cite{bonawitz2017practical,geyer2017differentially,hitaj2017deep,melis2019exploiting,zhu2019deep,mohri2019agnostic,wang2020federated}.  These studies are often under the implicit assumption that all learning phases during the training process is equally importantly.  Our work focuses on showing that the initial learning phase plays a critical role in the federated learning performance, which is orthogonal to the aforementioned studies.  

%% file: conclusion.tex
\section{Conclusion}

In this paper, we seized the existence of critical learning periods in federated learning so as to improve the federated learning training efficiency. Though a range of carefully designed experiments on different ML models and datasets, we showed that critical learning periods consistently exists in the training process of FL.  To explain such a phenomenon, we further proposed a new metric called Federated Fisher Information Matrix.  Our findings suggest that the initial learning phase plays a critical role in the final performance of FL.

%% file: appendix.tex
\section{Appendix}
In the appendix, we will provide our experiment details, parameter settings, and additional experimental results.  

\subsection{Dataset and Model}

To further observe critical learning periods in FL, we conduct extensive experiments with the CIFAR-10 dataset and the CIFAR-100 dataset \cite{krizhevsky2009learning} under both IID and Non-IID settings.  The CIFAR-10 dataset consists of $60,000$ $32\times32$ color images in $10$ classes, where $50,000$ samples are for training and the other $10,000$ samples for testing. Unlike CIFAR-10, CIFAR-100 has $100$ classes. We use the same strategy to distribute the data over different workers as suggested by \cite{mcmahan2017communication}. For the non-IID setting, we first divide each class of training data into ten parts, randomly assign three parts from different classes to each worker. For the IID setting, we evenly partition all training data among all workers.  
We consider two representative models: the ResNet-18 model \cite{he2016deep} and a five-layer CNN model, as shown in Table~\ref{tbl:resnet} and Table~\ref{tbl:cnn}. We run the experiments on PyTorch on Python3 with an NVIDIA RTX 3060 GPU. 

\begin{table}[!h]
  \centering
  \caption{The architecture of the ResNet-18 model.}
  \begin{tabular}{ccc}
    \toprule
    \textbf{Layer} & \textbf{Size} & \textbf{kernel\_size}\\ \otoprule
    conv1 & 64 & 3 \\[.05in]
    conv2.x & $\left[\begin{matrix} 3\times 3 & 64 \\ 3\times 3 & 64 \end{matrix}\right]\times 2$ & 3 \\[.15in]
    conv3.x & $\left[\begin{matrix} 3\times 3 & 128 \\ 3\times 3 & 128 \end{matrix}\right]\times 2$ & 3 \\[.15in]
    conv4.x & $\left[\begin{matrix} 3\times 3 & 256 \\ 3\times 3 & 256 \end{matrix}\right]\times 2$ & 3 \\[.15in]
    conv5.x & $\left[\begin{matrix} 3\times 3 & 512 \\ 3\times 3 & 512 \end{matrix}\right]\times 2$ & 3 \\[.15in]
    avg\_pool2d & 4 & - \\[.05in]
    Linear & 10 & - \\
    \bottomrule
  \end{tabular}

  \label{tbl:resnet}
\end{table}

\begin{table}[H]
  \centering
    \caption{The architecture of the five-layer CNN model.}
  \begin{tabular}{ccc}
    \toprule
    \textbf{Layer} & \textbf{Size} & \textbf{kernel\_size}\\ \otoprule
    conv2d & 64 & 3 \\ 
    conv2d & 192 & 3 \\
    conv2d & 384 & 3 \\
    conv2d & 256 & 3 \\
    conv2d & 256 & 3 \\
    Dropout & default & default \\
    Linear & 128 & - \\
    ReLU  & - & - \\
    Dropout & default & default \\
    Linear & 128 & - \\
    ReLU & - & - \\
    Linear & 10 & - \\
    \bottomrule
  \end{tabular}
  \label{tbl:cnn}
\end{table}

\begin{figure*}[t]
	\center
    	\includegraphics[width=0.95\textwidth]{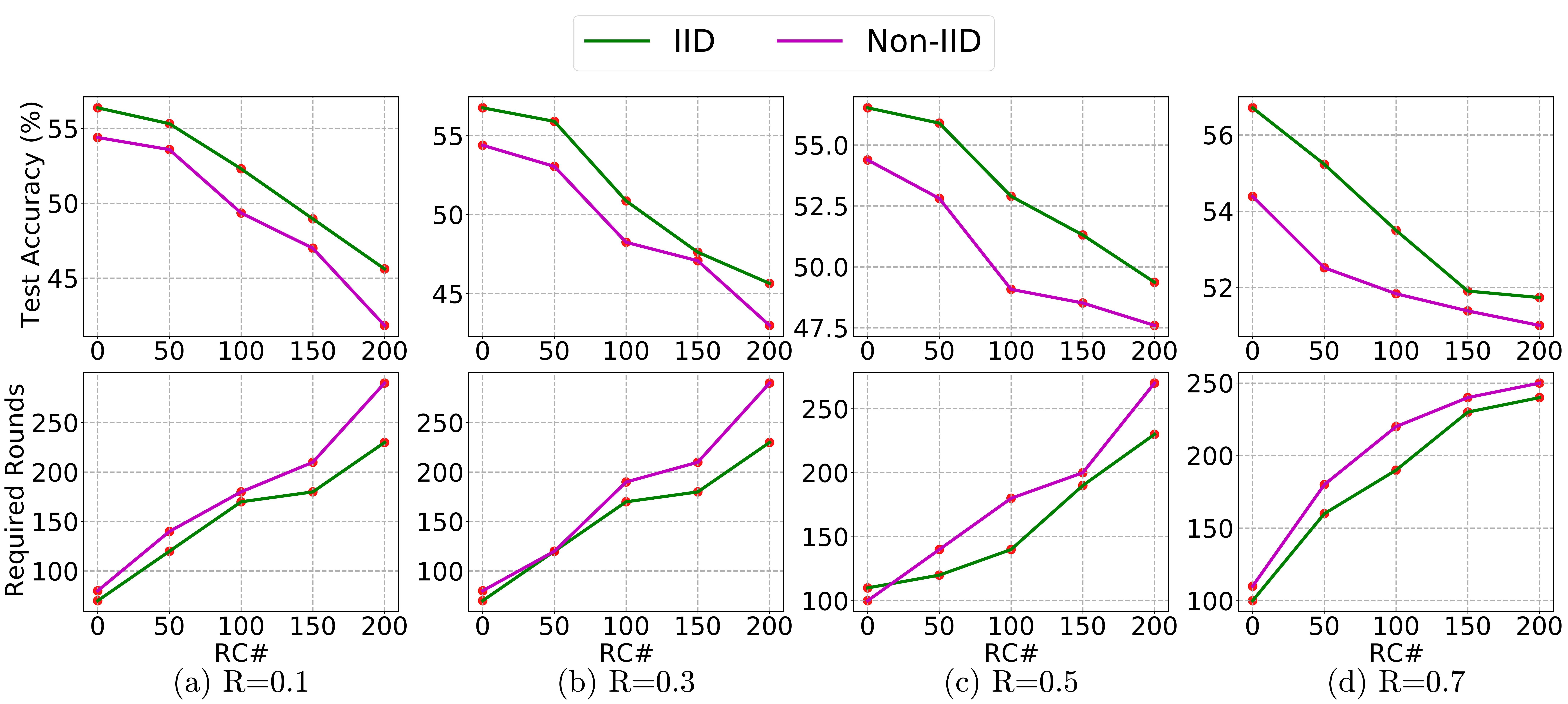}
		\caption{FL exhibits critical learning periods: \textbf{ResNet-18} on both IID and Non-IID \textbf{CIFAR-100} with FedAvg.  }
	\label{fig:acc-vs-recovery-all-cifar100}
\end{figure*}
\begin{figure*}[!t]
	\center
    	\includegraphics[width=0.95\textwidth]{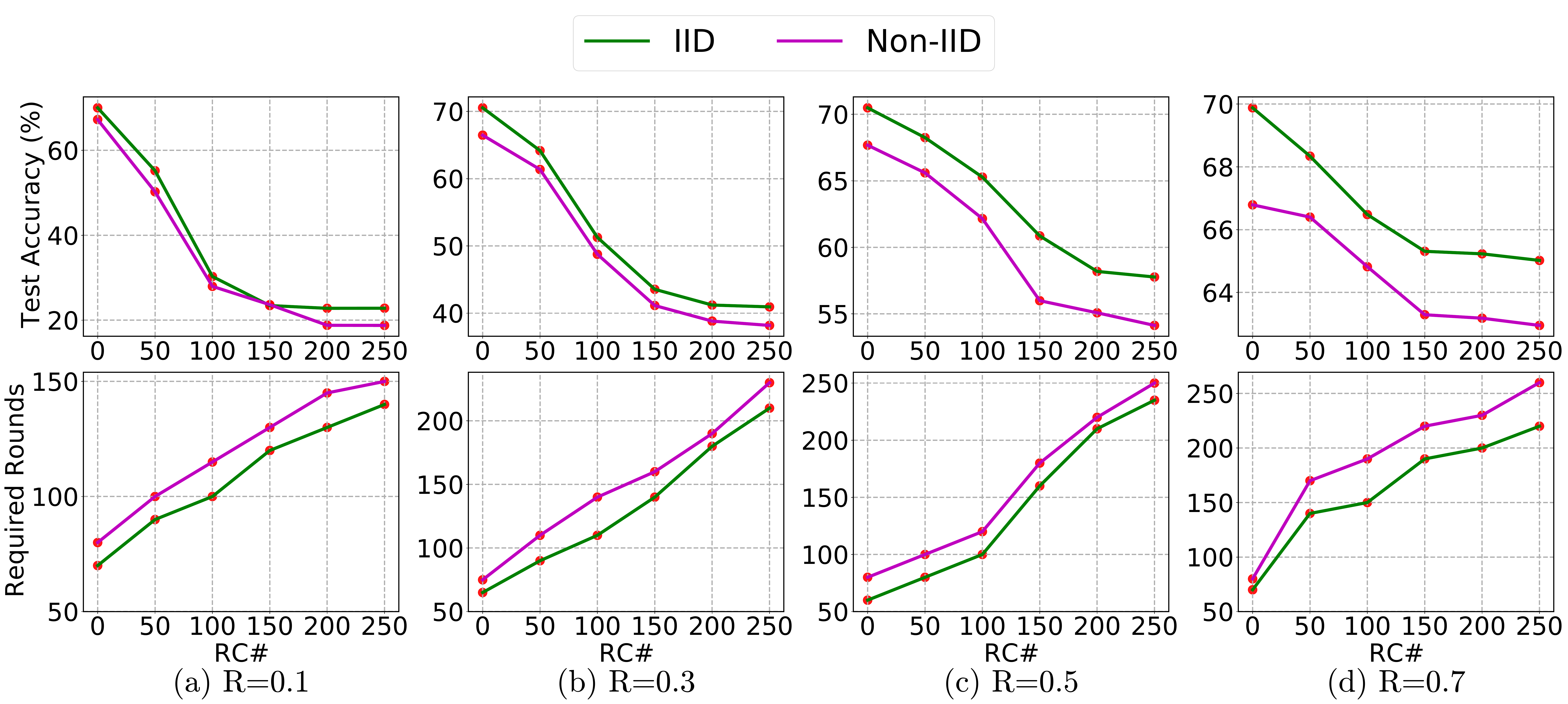}
		\caption{FL exhibits critical learning periods: \textbf{CNN} on both IID and Non-IID \textbf{CIFAR-10} with FedAvg. }
	\label{fig:acc-vs-recovery-all-cnn}
\end{figure*}

\begin{figure*}[h]
	\center
    	\includegraphics[width=0.95\textwidth]{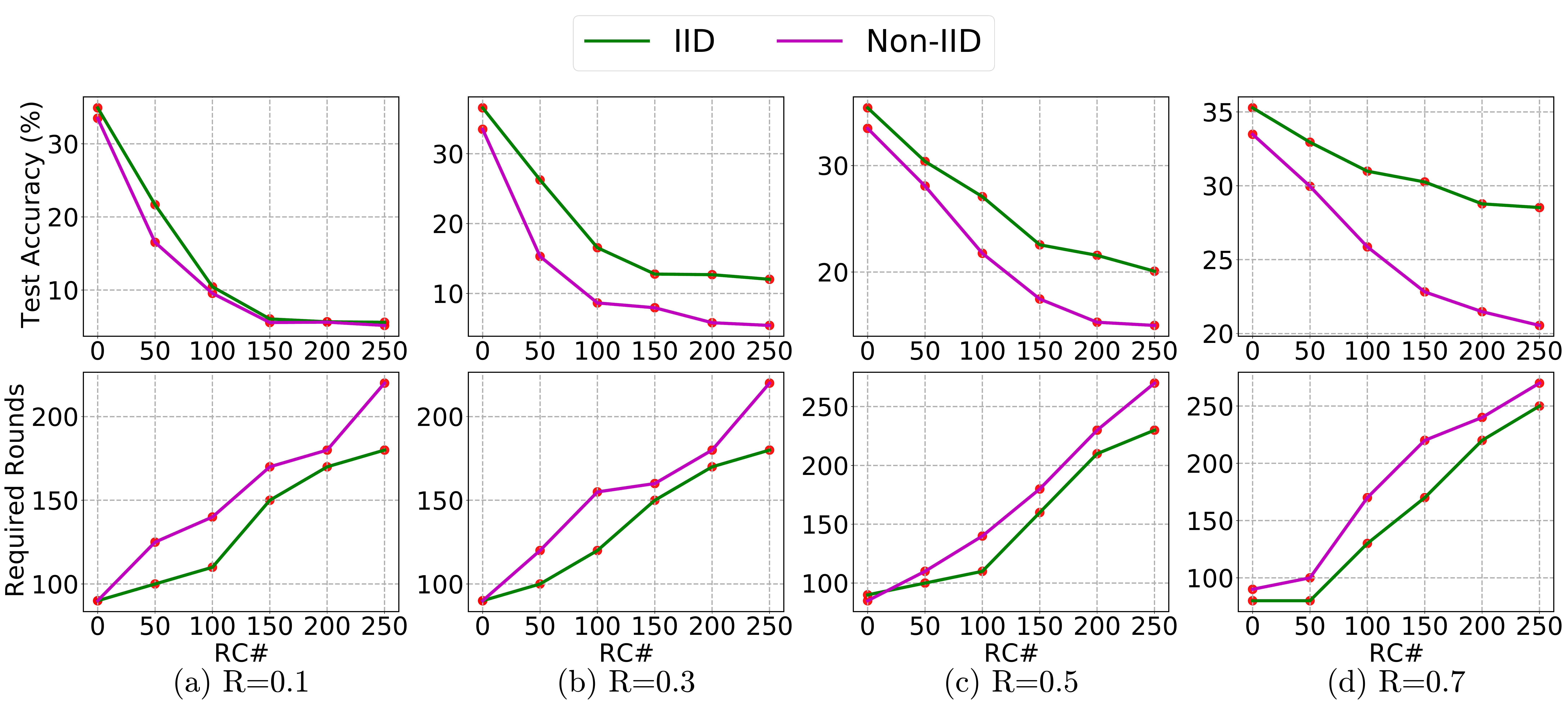}
		\caption{FL exhibits critical learning periods: \textbf{CNN} on both IID and Non-IID \textbf{CIFAR-100} with FedAvg.}
	\label{fig:acc-vs-recovery-all-cnn-cifar100}
\end{figure*}
\begin{figure*}[!t]
	\center
    	\includegraphics[width=0.99\textwidth]{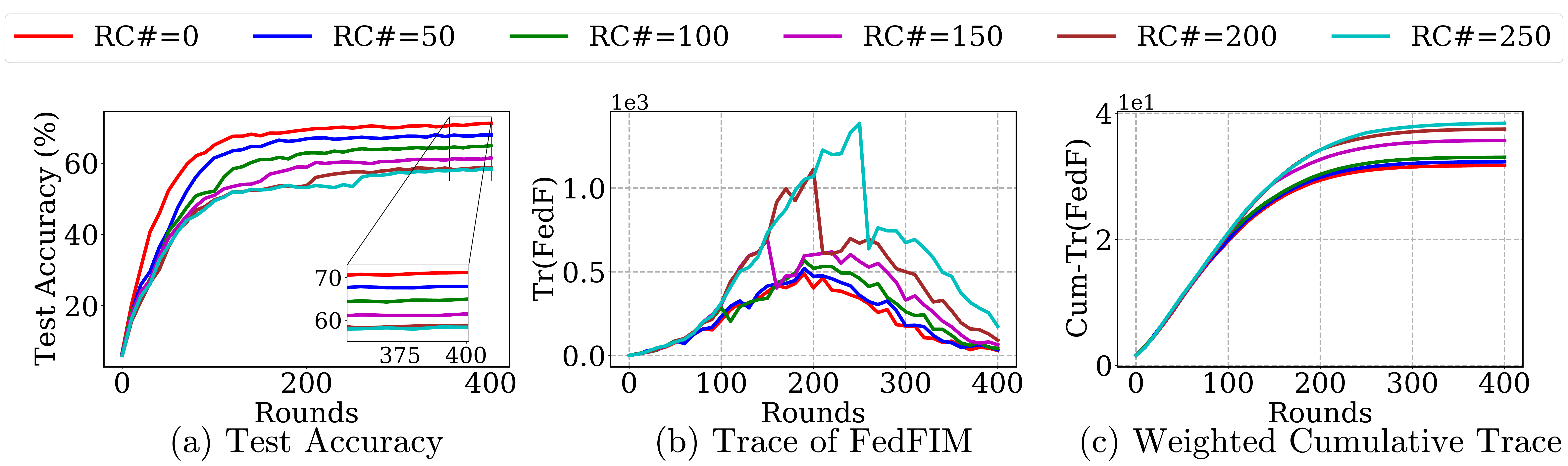}
		\caption{Connections between critical learning periods in FL and the Federated Fisher information achieved by \textbf{CNN} on IID \textbf{CIFAR-10} with FedAvg using 50\% of local datasets for training initially and recover to the entire datasets upon the recover round. (a) Test accuracy v.s. recover rounds: the final test accuracy is permanently impaired if the training dataset is not fully recovered at as early as the 20-th round. (b) Trace of FedFIM v.s. recover round. There exists a sharp increase of the trace of FedFIM in the early training phases. (c) Weighted cumulative sum of the trace of FedFIM vs. recover round. }
	\label{fig:fim-iid-cnn}
\end{figure*}

\subsection{Additional Results on Critical Learning Periods in FL}
Complementary to Section~\ref{sec:existence} ``Critical Learning Periods in FL'' on running the ResNet-18 model on CIFAR-10, we provide additional experimental results by running ResNet-18 on CIFAR-100, and CNN model on both CIFAR-10 and CIFAR-100 datasets.  We consider an FL system running FedAvg with $N=64$ clients that randomly selects a subset of $12$ clients in each round. The batch size is $16$, the initial learning rate is set to $0.01$ for ResNet-18 and $0.03$ for CNN with a decay of $0.97$ per round, and the SGD solver is adopted using an exponential annealing scheduling for the learning rate with a weight decay of $5\times 10^{-4}$.

We reproduce critical learning periods in FL when run ResNet-18 on CIFAR-100 under both IID and Non-IID settings, and CNN on CIFAR-10 and CIFAR-100 under both IID and Non-IID settings with FedAvg as shown in Figures~\ref{fig:acc-vs-recovery-all-cifar100},~\ref{fig:acc-vs-recovery-all-cnn},  and~\ref{fig:acc-vs-recovery-all-cnn-cifar100}, respectively. We observe that the critical learning periods consistently exist across all settings with different ratios of local datasets involved in the early learning phase. For example, if the CIFAR-10 training dataset is not recovered to the entire dataset at as early as the $20$--$50$-th communication rounds, the final test accuracy will be severely degraded compared to the standard FedAvg with the entire dataset.  Comparing among different ratios $R$ of local datasets involved in early training phase, it is not too surprising to see that a lower $R$ of local datasets in the early training phase makes drawing critical learning periods easier. 
Similarly, it is clear that the communication rounds are significantly increased with a lower final test accuracy as a function of the recover round $M$.  This further indicates the importance of the initial learning phase in determining the FL final performance.

\begin{figure*}
	\center
    	\includegraphics[width=0.99\textwidth]{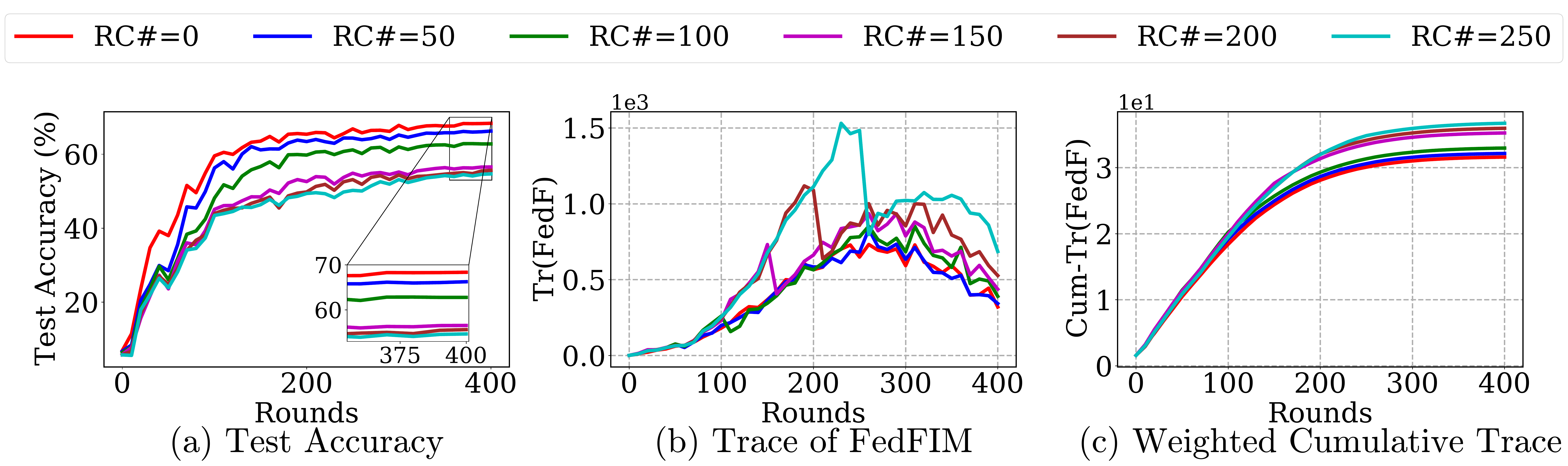}
		\caption{Connections between critical learning periods in FL and the Federated Fisher information achieved by \textbf{CNN} on Non-IID \textbf{CIFAR-10} with FedAvg using 50\% of local datasets for training initially and recover to the entire datasets upon the recover round.}
	\label{fig:fim-non-iid-cnn}
	\vspace{-0.1in}
\end{figure*}

\section{Additional Results on Federated Fisher Information}

Complementary to Section~\ref{sec:fim} ``Federated Fisher Information''  on running the ResNet-18 model on CIFAR-10, we provide additional experimental results by running the five-layer CNN model on CIFAR-10 and CIFAR-100.

\noindent\textbf{CNN on CIFAR-10.} We conduct similar experiments as in Figure~\ref{fig:acc-vs-recovery-all-cnn} with partial local datasets involved in the initial learning phases and the training datasets recover to the entire datasets at the ``recover rounds'' (RC\#).  Figures~\ref{fig:fim-iid-cnn} and~\ref{fig:fim-non-iid-cnn} present the test accuracy and the trace of FedFIM with different recover rounds when $R=0.5$, on IID and Non-IID CIFAR-10, respectively.  Again, we observe the existence of critical learning periods.  Furthermore, this information can be properly reflected via the trace of FedFIM as shown in Figure~\ref{fig:fim-iid-cnn}(b) for IID case and Figure~\ref{fig:fim-non-iid-cnn}(b) for Non-IID case.  There is a sharp increase in the trace of the FedFIM in the early phases of the FL training process, coinciding with a dramatic increase of the test accuracy in the early training phase.  The information starts to decrease when the test accuracy starts to plateau.  Since the training datasets are recovered from $50\%$ of local datasets to the entire datasets at the recover rounds, additional data further boosts the test accuracy as shown in Figure~\ref{fig:fim-iid-cnn}(b) and Figure~\ref{fig:fim-non-iid-cnn}(b).  However, such a test accuracy boost decreases significantly as the recover rounds increase.  This further suggests that the initial learning phases play a critical role in the FL and the permanent model degradation is irreversible no matter how much additional training is performed after the critical learning periods.  

We further consider the weighted cumulative sum of the trace of FedFIM.  The trace of FedFIM represents the degree of how good the local data is to improve the model---a larger values correspond to less model information---exactly shwon in Figure~\ref{fig:fim-iid-cnn} (c) and Figure~\ref{fig:fim-non-iid-cnn} (c), where a late recovery results in a larger weight cumulative trace.

\begin{figure*}
	\center
    	\includegraphics[width=0.99\textwidth]{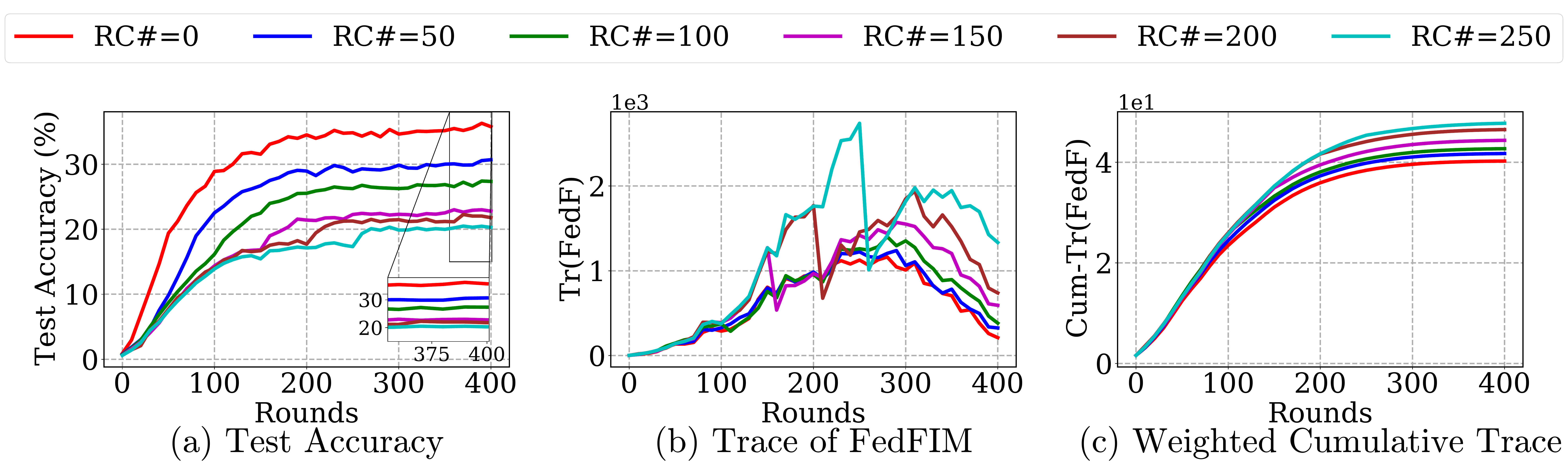}
		\caption{Connections between critical learning periods in FL and the Federated Fisher information achieved by \textbf{CNN} on IID \textbf{CIFAR-100} with FedAvg using 50\% of local datasets for training initially and recover to the entire datasets upon the recover round. }
	\label{fig:fim-iid-cnn-cifar100}
\end{figure*}

\begin{figure*}
	\center
    	\includegraphics[width=0.99\textwidth]{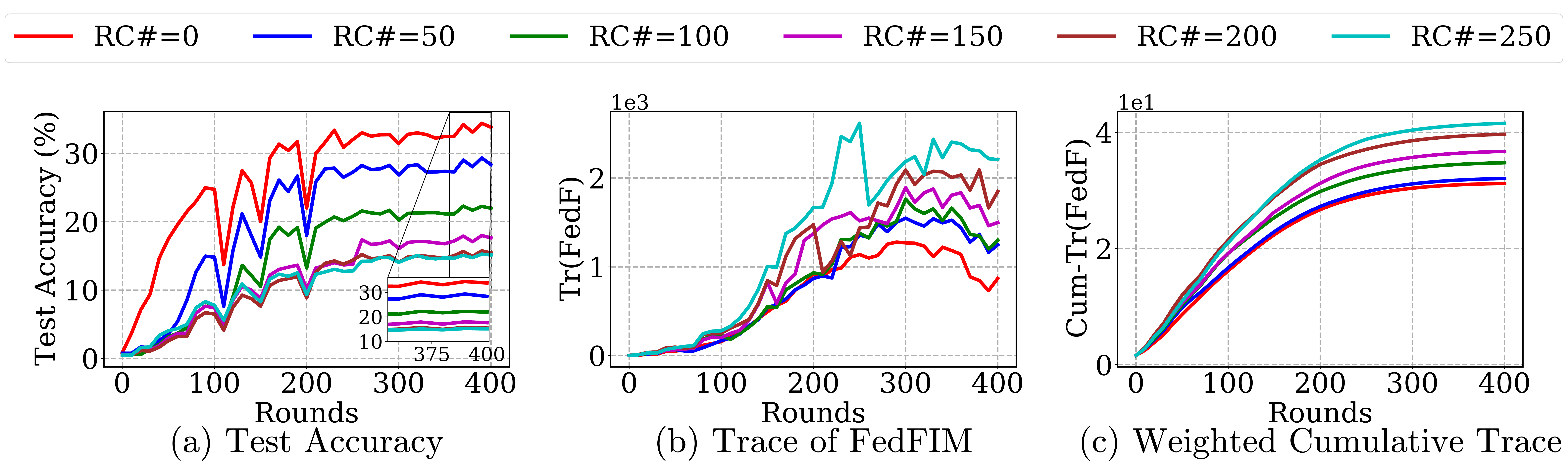}
		\caption{Connections between critical learning periods in FL and the Federated Fisher information achieved by \textbf{CNN} on Non-IID \textbf{CIFAR-100} with FedAvg using 50\% of local datasets for training initially and recover to the entire datasets upon the recover round.}
	\label{fig:fim-non-iid-cnn-cifar100}
	\vspace{-0.1in}
\end{figure*}

\noindent\textbf{CNN on CIFAR-100.} Similar observations and analysis apply to running CNN on CIFAR-100 as shown in Figures~\ref{fig:fim-iid-cnn-cifar100} and~\ref{fig:fim-non-iid-cnn-cifar100}, and hence are omitted here.

\begin{figure*}[!t]
	\center
    	\includegraphics[width=1\textwidth]{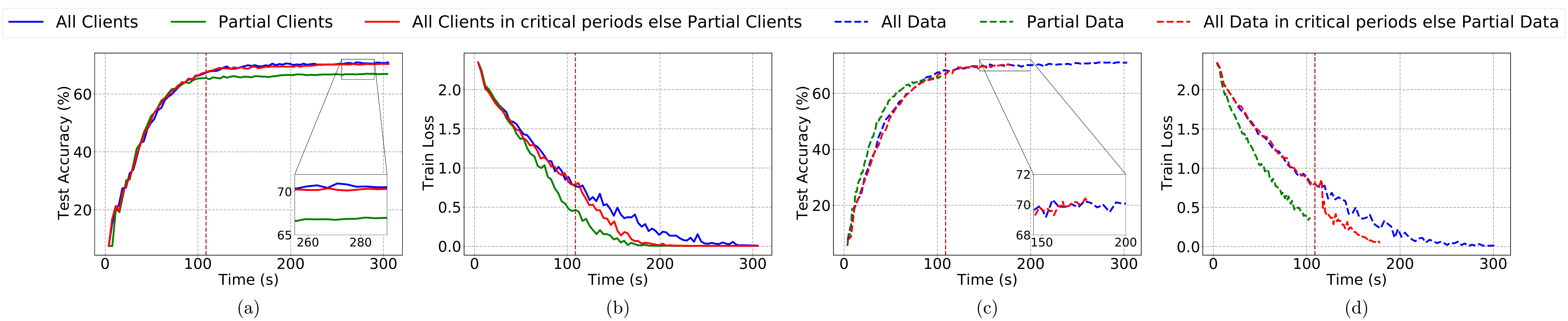}
    	\vspace{-0.2in}
		\caption{Seizing the critical learning periods in FL training with \textbf{CNN} on IID \textbf{CIFAR-10}.}
	\label{fig:motivation-iid-cnn}
\end{figure*}

\begin{figure*}[t]
	\center
    	\includegraphics[width=1\textwidth]{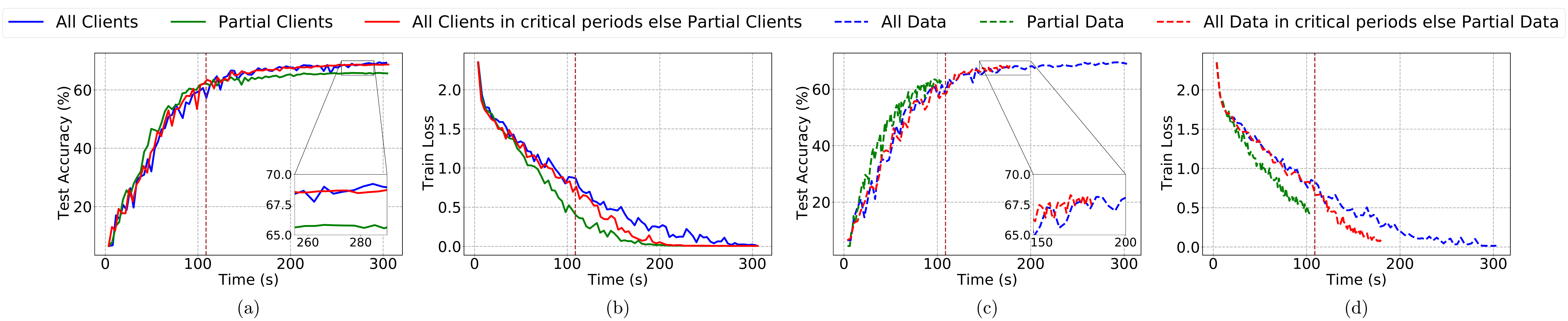}
    	\vspace{-0.2in}
		\caption{Seizing the critical learning periods in FL training with \textbf{CNN} on Non-IID \textbf{CIFAR-10}.}
	\label{fig:motivation-non-iid-cnn}
	\vspace{-0.1in}
\end{figure*}

\section{Additional Results on Seizing Critical Learning Periods}

Complementary to Section~\ref{sec:seizing} ``Seizing Critical Learning Periods,'' we further evaluate the idea that seizes critical learning periods to improve the FL training efficiency. Existing literature largely ignore the critical learning periods in FL training process.  In particular, we consider running the five-layer CNN on CIFAR-10. The total number of clients is $64$ and a subset of $12$ clients are randomly selected in each round. We consider the same settings as for ResNet-18 on CIFAR-10.

Figure~\ref{fig:motivation-iid-cnn}(a) and Figure~\ref{fig:motivation-non-iid-cnn}(a) show the test accuracy v.s. wall-clock time.  Again, we observe that there exists a requirement on the number of clients involved in training which provides similar test accuracy as using all clients (FedAvg) throughput.  For example, with all clients participating in the FL training during the critical learning periods, and then only 60\% of clients afterwards, the final test accuracy is similar to that using all clients throughout the training process.  Hence there is no need to involve all clients throughout the training process.  

Figure~\ref{fig:motivation-iid-cnn}(b) and Figure~\ref{fig:motivation-non-iid-cnn}(b) show the train loss v.s. wall-clock time. The participated client number requirement reduces the training time than using all clients (FedAvg) throughput. It is clear that leveraging critical learning periods for FL training, even in a heuristic manner, can significantly improve the training efficiency with a reduced training time while maintaining final test accuracy. 

Figure~\ref{fig:motivation-iid-cnn}(c) and Figure~\ref{fig:motivation-non-iid-cnn}(c) show the test accuracy v.s. wall-clock time. Again, we observe that there exists a training dataset requirement which provides similar test accuracy as using the entire dataset (FedAvg) throughput.  For example, with the entire training datasets used in the FL training during the critical learning periods, and then only 25\% of local datasets afterwards, the final test accuracy is similar to that using the entire datasets throughout the training process. Hence, there is no need to use the entire training datasets throughout the training process.  
Figure~\ref{fig:motivation-iid-cnn}(d) and Figure~\ref{fig:motivation-non-iid-cnn}(d) present the train loss v.s. wall-clock time, the training dataset requirement (the heuristic) reduces the training time than using the entire dataset (FedAvg) throughput.  Again, we observe that the early learning phase plays a critical role in FL performance, and leveraging it can significantly improve the training efficiency of FL.

%% file: FLcriticalperiodsarXiv.bbl
\begin{thebibliography}{10}

\bibitem{mcmahan2017communication}
Brendan McMahan, Eider Moore, Daniel Ramage, Seth Hampson, and Blaise~Aguera
  y~Arcas.
\newblock {Communication-Efficient Learning of Deep Networks from Decentralized
  Data}.
\newblock In {\em Proc. of AISTATS}, 2017.

\bibitem{kairouz2019advances}
Peter Kairouz, H~Brendan McMahan, Brendan Avent, Aur{\'e}lien Bellet, Mehdi
  Bennis, Arjun~Nitin Bhagoji, Kallista Bonawitz, Zachary Charles, Graham
  Cormode, Rachel Cummings, et~al.
\newblock {Advances and Open Problems in Federated Learning}.
\newblock {\em arXiv preprint arXiv:1912.04977}, 2019.

\bibitem{imteaj2021survey}
Ahmed Imteaj, Urmish Thakker, Shiqiang Wang, Jian Li, and M~Hadi Amini.
\newblock {A Survey on Federated Learning for Resource-Constrained IoT
  Devices}.
\newblock {\em IEEE Internet of Things Journal}, 2021.

\bibitem{yang2018applied}
Timothy Yang, Galen Andrew, Hubert Eichner, Haicheng Sun, Wei Li, Nicholas
  Kong, Daniel Ramage, and Fran{\c{c}}oise Beaufays.
\newblock {Applied Federated Learning: Improving Google Keyboard Query
  Suggestions}.
\newblock {\em arXiv preprint arXiv:1812.02903}, 2018.

\bibitem{konevcny2016federated}
Jakub Kone{\v{c}}n{\`y}, H~Brendan McMahan, Felix~X Yu, Peter Richt{\'a}rik,
  Ananda~Theertha Suresh, and Dave Bacon.
\newblock {Federated Learning: Strategies for Improving Communication
  Efficiency}.
\newblock {\em arXiv preprint arXiv:1610.05492}, 2016.

\bibitem{suresh2017distributed}
Ananda~Theertha Suresh, X~Yu Felix, Sanjiv Kumar, and H~Brendan McMahan.
\newblock {Distributed Mean Estimation with Limited Communication}.
\newblock In {\em Proc. of ICML}, 2017.

\bibitem{caldas2018expanding}
Sebastian Caldas, Jakub Kone{\v{c}}ny, H~Brendan McMahan, and Ameet Talwalkar.
\newblock {Expanding the Reach of Federated Learning by Reducing Client
  Resource Requirements}.
\newblock {\em arXiv preprint arXiv:1812.07210}, 2018.

\bibitem{xu2019elfish}
Zirui Xu, Zhao Yang, Jinjun Xiong, Jianlei Yang, and Xiang Chen.
\newblock {ELFISH: Resource-Aware Federated Learning on Heterogeneous Edge
  Devices}.
\newblock {\em arXiv preprint arXiv:1912.01684}, 2019.

\bibitem{wang2019adaptivecom}
Jianyu Wang and Gauri Joshi.
\newblock {Adaptive Communication Strategies to Achieve the Best Error-Runtime
  Trade-off in Local-update SGD}.
\newblock In {\em Proc. of SysML}, 2019.

\bibitem{wang2019adaptive}
Shiqiang Wang, Tiffany Tuor, Theodoros Salonidis, Kin~K Leung, Christian
  Makaya, Ting He, and Kevin Chan.
\newblock {Adaptive Federated Learning in Resource Constrained Edge Computing
  Systems}.
\newblock {\em IEEE Journal on Selected Areas in Communications},
  37(6):1205--1221, 2019.

\bibitem{karimireddy2020scaffold}
Sai~Praneeth Karimireddy, Satyen Kale, Mehryar Mohri, Sashank Reddi, Sebastian
  Stich, and Ananda~Theertha Suresh.
\newblock {SCAFFOLD: Stochastic Controlled Averaging for Federated Learning}.
\newblock In {\em Proc. of ICML}, 2020.

\bibitem{lai2021oort}
Fan Lai, Xiangfeng Zhu, Harsha~V Madhyastha, and Mosharaf Chowdhury.
\newblock {Oort: Efficient Federated Learning via Guided Participant
  Selection}.
\newblock In {\em Proc. of USENIX OSDI}, 2021.

\bibitem{wang2020optimizing}
Hao Wang, Zakhary Kaplan, Di~Niu, and Baochun Li.
\newblock {Optimizing Federated Learning on Non-IID Data With Reinforcement
  Learning}.
\newblock In {\em Proc. of IEEE INFOCOM}, 2020.

\bibitem{xiong2021straggler}
Guojun Xiong, Gang Yan, and Jian Li.
\newblock {Straggler-Resilient Distributed Machine Learning with Dynamic Backup
  Workers}.
\newblock {\em arXiv preprint arXiv:2102.06280}, 2021.

\bibitem{achille2019critical}
Alessandro Achille, Matteo Rovere, and Stefano Soatto.
\newblock {Critical Learning Periods in Deep Networks}.
\newblock In {\em Proc. of ICLR}, 2019.

\bibitem{jastrzebski2019relation}
Stanislaw Jastrzebski, Zachary Kenton, Nicolas Ballas, Asja Fischer, Yoshua
  Bengio, and Amos~J Storkey.
\newblock {On the Relation Between the Sharpest Directions of DNN Loss and the
  SGD Step Length}.
\newblock In {\em Proc. of ICLR}, 2019.

\bibitem{golatkar2019time}
Aditya~Sharad Golatkar, Alessandro Achille, and Stefano Soatto.
\newblock {Time Matters in Regularizing Deep Networks: Weight Decay and Data
  Augmentation Affect Early Learning Dynamics, Matter Little Near Convergence}.
\newblock {\em Proc. of NeurIPS}, 2019.

\bibitem{jastrzebski2021catastrophic}
Stanislaw Jastrzebski, Devansh Arpit, Oliver Astrand, Giancarlo~B Kerg, Huan
  Wang, Caiming Xiong, Richard Socher, Kyunghyun Cho, and Krzysztof~J Geras.
\newblock {Catastrophic Fisher Explosion: Early Phase Fisher Matrix Impacts
  Generalization}.
\newblock In {\em Proc. of ICML}, 2021.

\bibitem{amari2000methods}
Shun-ichi Amari and Hiroshi Nagaoka.
\newblock {\em {Methods of Information Geometry}}, volume 191.
\newblock American Mathematical Soc., 2000.

\bibitem{li2020federated}
Tian Li, Anit~Kumar Sahu, Manzil Zaheer, Maziar Sanjabi, Ameet Talwalkar, and
  Virginia Smith.
\newblock {Federated Optimization in Heterogeneous Networks}.
\newblock In {\em Proc. of MLSys}, 2020.

\bibitem{rothchild2020fetchsgd}
Daniel Rothchild, Ashwinee Panda, Enayat Ullah, Nikita Ivkin, Ion Stoica,
  Vladimir Braverman, Joseph Gonzalez, and Raman Arora.
\newblock {FetchSGD: Communication-Efficient Federated Learning with
  Sketching}.
\newblock In {\em Proc. of ICML}, 2020.

\bibitem{agarwal2020accordion}
Saurabh Agarwal, Hongyi Wang, Kangwook Lee, Shivaram Venkataraman, and Dimitris
  Papailiopoulos.
\newblock {ACCORDION: Adaptive Gradient Communication via Critical Learning
  Regime Identification}.
\newblock In {\em Proc. of MLSys}, 2021.

\bibitem{he2016deep}
Kaiming He, Xiangyu Zhang, Shaoqing Ren, and Jian Sun.
\newblock {Deep Residual Learning for Image Recognition}.
\newblock In {\em Proc. of IEEE CVPR}, 2016.

\bibitem{krizhevsky2009learning}
Alex Krizhevsky, Geoffrey Hinton, et~al.
\newblock {Learning Multiple Layers of Features from Tiny Images}.
\newblock 2009.

\bibitem{martens2014new}
James Martens.
\newblock {New Insights and Perspectives on the Natural Gradient Method}.
\newblock {\em arXiv preprint arXiv:1412.1193}, 2014.

\bibitem{jastrzkebski2017three}
Stanis{\l}aw Jastrz{\k{e}}bski, Zachary Kenton, Devansh Arpit, Nicolas Ballas,
  Asja Fischer, Yoshua Bengio, and Amos Storkey.
\newblock {Three Factors Influencing Minima in SGD}.
\newblock {\em arXiv preprint arXiv:1711.04623}, 2017.

\bibitem{smith2018don}
Samuel~L Smith, Pieter-Jan Kindermans, Chris Ying, and Quoc~V Le.
\newblock {Don't Decay the Learning Rate, Increase the Batch Size}.
\newblock In {\em Proc. of ICLR}, 2018.

\bibitem{smith2017federated}
Virginia Smith, Chao-Kai Chiang, Maziar Sanjabi, and Ameet Talwalkar.
\newblock {Federated Multi-Task Learning}.
\newblock In {\em Proc. of NIPS}, 2017.

\bibitem{li2020convergence}
Xiang Li, Kaixuan Huang, Wenhao Yang, Shusen Wang, and Zhihua Zhang.
\newblock {On the Convergence of FedAvg on Non-IID Data}.
\newblock In {\em Proc. of ICLR}, 2020.

\bibitem{liu2020federated}
Fenglin Liu, Xian Wu, Shen Ge, Wei Fan, and Yuexian Zou.
\newblock {Federated Learning for Vision-and-Language Grounding Problems}.
\newblock In {\em Proc. of AAAI}, 2020.

\bibitem{wang2020federated}
Hongyi Wang, Mikhail Yurochkin, Yuekai Sun, Dimitris Papailiopoulos, and
  Yasaman Khazaeni.
\newblock {Federated Learning with Matched Averaging}.
\newblock In {\em Proc. of ICLR}, 2020.

\bibitem{bonawitz2017practical}
Keith Bonawitz, Vladimir Ivanov, Ben Kreuter, Antonio Marcedone, H~Brendan
  McMahan, Sarvar Patel, Daniel Ramage, Aaron Segal, and Karn Seth.
\newblock {Practical Secure Aggregation for Privacy-Preserving Machine
  Learning}.
\newblock In {\em Proc. of ACM CCS}, 2017.

\bibitem{geyer2017differentially}
Robin~C Geyer, Tassilo Klein, and Moin Nabi.
\newblock {Differentially Private Federated Learning: A Client Level
  Perspective}.
\newblock {\em arXiv preprint arXiv:1712.07557}, 2017.

\bibitem{hitaj2017deep}
Briland Hitaj, Giuseppe Ateniese, and Fernando Perez-Cruz.
\newblock {Deep Models under the GAN: Information Leakage from Collaborative
  Deep Learning}.
\newblock In {\em Proc. of ACM CCS}, 2017.

\bibitem{melis2019exploiting}
Luca Melis, Congzheng Song, Emiliano De~Cristofaro, and Vitaly Shmatikov.
\newblock {Exploiting Unintended Feature Leakage in Collaborative Learning}.
\newblock In {\em Proc. of IEEE S\&P}, 2019.

\bibitem{zhu2019deep}
Ligeng Zhu, Zhijian Liu, and Song Han.
\newblock {Deep Leakage from Gradients}.
\newblock {\em Proc. of NeurIPS}, 2019.

\bibitem{mohri2019agnostic}
Mehryar Mohri, Gary Sivek, and Ananda~Theertha Suresh.
\newblock {Agnostic Federated Learning}.
\newblock In {\em Proc. of ICML}, 2019.

\end{thebibliography}
